# GateFuseNet: An Adaptive 3D Multimodal Neuroimaging Fusion Network for Parkinson's Disease Diagnosis

Rui Jin[1, #], Chen Chen[1, #], Yin Liu[3], Hongfu Sun[2], Min Zeng[1], Min Li[1], Yang Gao[1, *]

[1]School of Computer Science and Engineering, Central South University, Changsha, 410083, China

[2]School of Engineering, University of Newcastle, Newcastle, Callaghan, New South Wales 2308, Australia

[3]Department of Radiology, The Third Xiangya Hospital, Central South University, Changsha, China

[#] These authors contributed equally to this work

* Correspondence to: Yang Gao, E-mail: yang.gao@csu.edu.cn

*Abstract*—Accurate diagnosis of Parkinson's disease (PD) from MRI remains challenging due to symptom variability and pathological heterogeneity. Most existing methods rely on conventional magnitude-based MRI modalities, such as T1-weighted images (T1w), which are less sensitive to PD pathology than Quantitative Susceptibility Mapping (QSM), a phase-based MRI technique that quantifies iron deposition in deep gray matter nuclei. In this study, we propose GateFuseNet, an adaptive 3D multimodal fusion network that integrates QSM and T1w images for PD diagnosis. The core innovation lies in a gated fusion module that learns modality-specific attention weights and channel-wise gating vectors for selective feature modulation. This hierarchical gating mechanism enhances ROI-aware features while suppressing irrelevant signals. Experimental results show that our method outperforms three existing state-of-the-art approaches, achieving 85.00% accuracy and 92.06% AUC. Ablation studies further validate the contributions of ROI guidance, multimodal integration, and fusion positioning. Grad-CAM visualizations confirm the model's focus on clinically relevant pathological regions. The source codes and pretrained models can be found at https://github.com/YangGaoUQ/GateFuseNet

Keywords—Parkinson's Disease, QSM, MRI, Multi-Neuroimaging Modalities, GateFuseNet

## I. Introduction

Parkinson's disease (PD) is a progressive neurodegenerative disorder characterized by dopaminergic neuron loss in the substantia nigra (SN) and pathological iron accumulation in deep gray matter (DGM) nuclei. These pathological changes contribute to the core motor symptoms of tremor, rigidity, and bradykinesia [1]. Current clinical assessments, such as the Unified Parkinson's Disease Rating Scale (UPDRS), remain largely subjective and lack sensitivity to early-stage pathology [2].

Magnetic resonance imaging (MRI) provides a non-invasive means to investigate PD-related neurodegeneration [3]. However, conventional modalities (e.g., T1- and T2-weighted imaging) offer only qualitative information and cannot quantify iron deposition, a key pathological hallmark of PD [4]. Quantitative Susceptibility Mapping (QSM) enables in vivo assessment of tissue magnetic susceptibility, reflecting iron content, and has shown great promise as a biomarker for PD [5, 6]. Yet, QSM often exhibits boundary ambiguity in certain DGM regions (e.g., caudate nucleus, putamen), while T1-weighted imaging (T1WI) provides superior anatomical definition [7]. Integrating QSM-derived pathological contrast with T1WI anatomical precision can therefore enhance both sensitivity and interpretability in PD diagnosis. Despite their complementary advantages, only limited efforts have been made to jointly exploit QSM and T1WI for PD diagnosis, and most existing works treat them independently.

Deep learning (DL) has achieved notable success in MRI-based PD research due to its ability to automatically extract discriminative imaging features [8]. However, most existing approaches rely on single modalities and generic architectures, limiting their ability to capture complementary structural and pathological cues. Furthermore, current models often neglect domain-specific anatomical priors that could guide feature learning toward clinically relevant regions. Moreover, conventional CNNs and transformers lack anatomically guided attention, and their "black-box" nature impedes clinical interpretability [9, 10]. Recent attempts at multimodal fusion or attention-based designs still face challenges such as reliance on pre-defined ROIs, limited interpretability, or high computational cost [11]. In addition, although explainable DL techniques and biologically inspired architectures have emerged, their integration into PD imaging remains in its infancy. Future multimodal frameworks that jointly leverage data-driven representations and anatomical priors may provide a more robust and interpretable pathway for precision diagnosis.

To address these gaps, we propose GateFuseNet, a unified, anatomically informed multimodal framework that adaptively integrates QSM and T1WI for PD diagnosis. Deep gray matter ROI masks are explicitly incorporated as anatomical guidance. GateFuseNet employs modality-specific 3D encoders enhanced with Convolutional Block Attention Modules (CBAMs) and Gated Fusion (GF) blocks for adaptive attention-based integration. This design selectively enhances disease-relevant features while suppressing irrelevant signals.

Extensive experiments demonstrate GateFuseNet's superior diagnostic accuracy and anatomical interpretability, with Grad-CAM visualizations confirming its consistent focus on PD-relevant nuclei.

## II. MATERIALS AND METHODS

*A. Subject Recruitment and Image Acquisition*

A total of 316 subjects were recruited from September, 2023 to September, 2024, comprising 155 healthy controls (HC) and 161 individuals with Parkinson's disease (PD). Among these participants, 64 subjects were randomly spared as an independent test cohort for subsequent comparative studies and ablation experiments, leaving 252 subjects for model training. All MRI examinations were performed on a 3T scanner (vendor/model to be specified) equipped with a 32-channel phased-array head coil. Each participant underwent two imaging protocols: a 3D turbo fast-echo (3D-TFE) T1-weighted sequence and a 3D multi-echo gradient-echo (GRE) sequence, detailed as follows:

**QSM Acquisition:** The 3D multi-echo GRE sequence was performed with the following parameters: TR = 35 ms; TE = 4.1 ms with a delta TE of 4.0 ms; FOV = 196 × 196 × 128 mm³; reconstruction voxel size = 0.88 × 0.88 × 1.00 mm³; flip angle = 20°; acquisition mode = Cartesian; flow compensation = applied; total acquisition time = approximately 5.43 minutes; SENSE acceleration factor = 3. All QSM images were reconstructed using a recently proposed iQSM+ [12] method, which can directly produce high-quality QSM images from the MRI raw phases in a single step.

**T1w Images Acquisition:** MRI scans were acquired using a Philips 3.0T scanner (Ingenia Elition X; Philips Medical Systems, Best, the Netherlands) with a 32-channel phased-array head coil. Sagittal T1 structural images were acquired using the three-dimensional turbo fast echo (3D-TFE) T1WI sequence to capture high resolution images. The parameters were set as follows: repetition time (TR)/echo time (TE) =8.3/3.8 ms; flip angle =8°; turbo factor = 240; field of view (FOV) = 240 × 240 × 170 mm3; voxel size = 0.94 × 0.94 × 1.0 mm; acquisition matrix = 256 × 256 × 170; compressed sensing acceleration factor = 3; total acquisition time = 3 min 5 s.

*B. DGM ROI segmentation*

An atlas-based segmentation pipeline was conducted to produce DGM ROI parcellation maps for each subject. First, the QSM images of all participants were co-registered to the QSM-specific MuSus-100 template in MNI space [13] to obtain forward transformation and inverse transformation parameters using Advanced Normalization Tools (ANTs) [14]. A total of 170 ROIs covering the whole brain were then extracted by registering a customized atlas constructed through non-linear registration of the Automated Anatomical Labeling Atlas (AAL3) [15] to the MuSus-100 template via the HybraPD protocol [16] back to all individual images. We specifically selected several DGM regions, including bilateral substantia nigra, putamen, caudate nucleus, globus pallidus, and Subthalamic nucleus as the primary ROI inputs to our network, since they have been reported to be important for PD diagnosis [17]. All registered QSM images and segmentation masks underwent independent reviews by two radiologists to exclude cases with significant misregistration or segmentation errors.

*C. Overview of model architecture*

Fig. 1 shows the overall framework of the proposed GateFuseNet, which mainly comprises three branches, taking QSM, T1w images and pathological-aware ROI parcellation maps as inputs, respectively. The network consists of a Stem Module, followed by a GF block, then three Fusion Modules, and a final Decision Module.

Module processes each modality through modality-specific encoders. After the Stem, a GF block is applied to fuse the features from all three input branches. Each Fusion Module consists of three parallel branches, processed independently through three CBAM-augmented [18] bottleneck blocks, followed by a residual fusing block that employs a GF block to effectively combine the features. The Decision Module consists of three CBAM-dilated bottleneck blocks followed by a fully connected layer for classification. This architecture supports progressive abstraction of multimodal features through hierarchical fusion while preserving modality-specific information and leveraging their complementary diagnostic cues.

*D. Stem Module*

GateFuseNet begins with three parallel 3D encoders that are customized to the signal characteristics of each modality while producing shape-compatible feature maps for subsequent fusion (Fig. 1 (b)). This design ensures that low-level features are preserved while reducing computational complexity.

The Stem Module serves as the initial feature extractor for each input modality containing QSM, T1w image, and pathological-aware ROI masks. Each modality is processed by a modality-specific 3D encoder consisting of three stacked 3×3×3 convolutional layers with ELU activations, followed by a 2×2×2 max-pooling layer for downsampling. This module transforms the input volumes into shape-compatible feature representations, preserving modality-specific characteristics while standardizing spatial dimensions for subsequent fusion. The outputs from the three stems are passed to the first Gated Fusion (GF) block for adaptive multimodal integration.

*E. Fusion Module*

Each Fusion Module in GateFuseNet is designed to progressively refine and integrate multimodal features in a pathology-aware manner. It consists of three parallel branches, responsible for modality-specific representation learning: QSM information extraction branch, T1w anatomical feature branch, and pathological-aware ROI guidance branch.

Following independent encoding, the outputs from the three branches are passed to a GF block, the core component of Fusion Module, which performs adaptive feature integration. Specifically, the Adaptive Multimodal Fusion (AMF) component within the GF block learns voxel-wise attention weights for each modality, while the channel-wise gating (CWG) mechanism determines how much of the fused signal is injected into the pathological-aware ROI guidance path, enabling the network to suppress redundant or noisy information and emphasize disease-relevant signals. This design ensures that the fusion process is both anatomically informed and dynamically adaptive across different feature hierarchies.

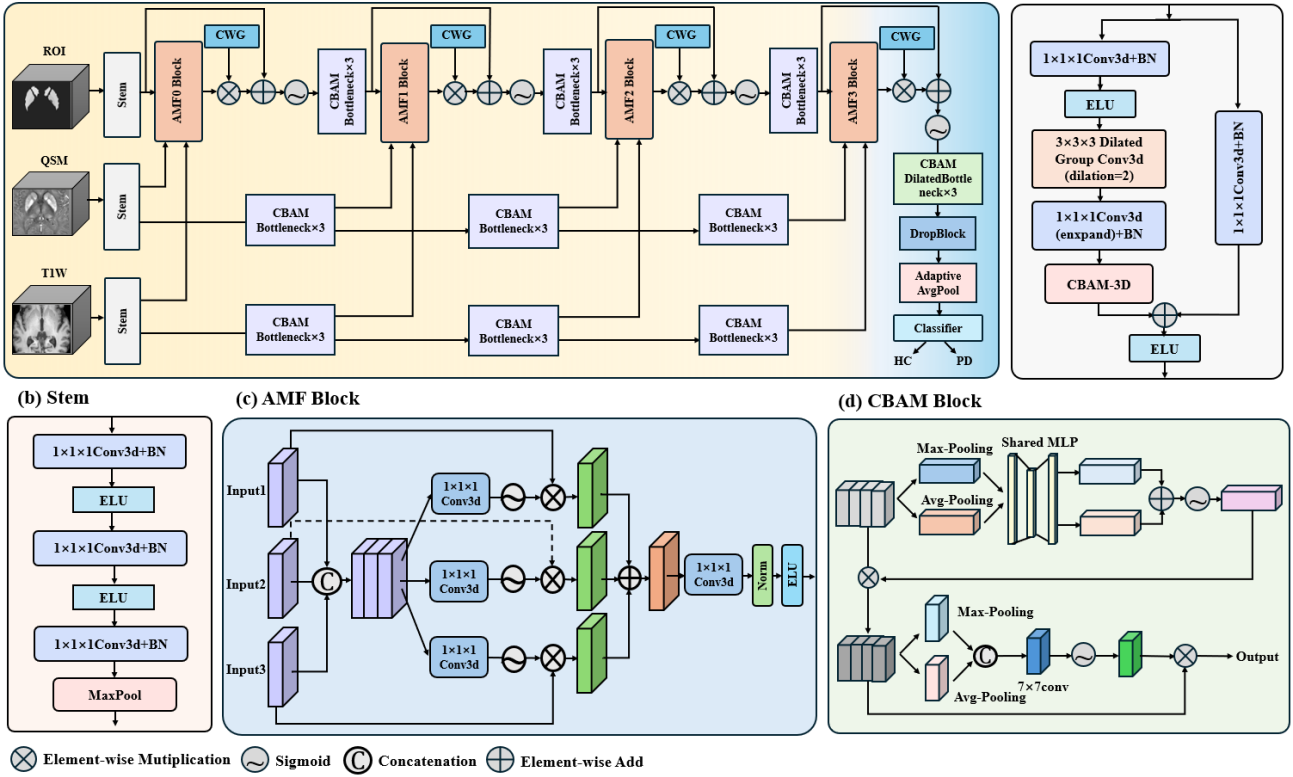

Fig. 1 Overall framework of the proposed GateFuseNet for diagnosing PD, taking QSM and T1w images as well as ROI parcellation maps of deep brain nucleus as its inputs. The proposed network is mainly composed the following modules: (b) Stem module for initial feature extraction, (c) AMF block for producing modality-specific attention weights for adaptive multi-modal latent feature fusion, (d) CBAM block to combine channel and spatial attention mechanisms to refine latent features, and (e) CBAMDilatedBottleneck with grouped dilated convolutions and residual connections for final feature aggregation.

*F. AMF Block and CWG*

Given features $x_l^m \in R^{C \times D \times H \times W}$, for modality $m \in \{ROI, QSM, T1\}$ at stage $\ell$, the GF block proceeds in two cascaded steps.

Fig. 1 (c) depicts the architecture of the AMF block. The three tensors are concatenated and fed to three parallel $3 \times 3 \times 3$ grouped-convolution heads that generate modality-specific attention maps:

$$\alpha_l^m = \sigma\left(BN\left(Conv_{3\times3\times3}^m[x_l^{ROI}, x_l^{qsm}, x_l^{T1}]\right)\right),$$
$$m \in \{ROI, QSM, T1\}, \quad (1)$$

where $\sigma$ denotes the sigmoid. A voxel-wise normalization enforces a convex combination, ensuring the three weights sum to 1 at every voxel,

$$\widetilde{\alpha}_l^m = \frac{\alpha_l^m}{\sum_n \alpha_l^n + \varepsilon} \quad (2)$$

and yields the fused feature tensor

$$f_l = \sum_m \widetilde{\alpha}_l^m \odot \alpha_l^m. \quad (3)$$

Thus, a modal level attention mechanism is implemented to determine which modality should dominate each spatial position.

While AMF decides which modality dominates at each voxel, it does not regulate how much of the fused tensor enters the residual stream. We therefore introduce a learnable gate vector $\theta_l \in R^C$, where each element corresponds to one feature channel. After a sigmoid squash,

$$\widehat{f_l} = \sigma(\theta_l) \odot f_l, \quad (4)$$

and $g_l$ is broadcast over the $D \times H \times W$. Channels incompatible with the current ROI context are thus suppressed, whereas informative ones are retained before residual injection.

The gated tensor is injected into the hybrid branch guided by ROI information via a residual addition,

$$x_{l+1}^{ROI} = x_l^{ROI} + \widehat{f_l}, \quad (5)$$

while $x_l^{QSM}$ and $x_l^{T1}$ propagate unchanged. Deployed at every stage, this block progressively integrates multimodal features while maintaining fine-grained control over their per-channel contributions.

*G. Decision Module*

The Decision Module is the final stage in GateFuseNet, where high-level semantic reasoning is performed for classification. As shown in Fig. 1 (e), this module consists of three CBAM-augmented dilated bottleneck blocks. Each block first applies 1x1 convolutions followed by a grouped 3x3 convolution with dilations, expanding the receptive field while maintaining spatial resolution. The use of CBAM (Convolutional Block Attention Module) further enhances feature representation by focusing attention on important regions, which is crucial for detecting subtle pathological cues in Parkinson's disease. After passing through these blocks, a global average pooling layer is applied to reduce feature dimensions, followed by a fully connected layer that generates the final classification output. This structure ensures that the model captures both local and contextual information, providing high precision in PD diagnosis while maintaining anatomical specificity.

*H. Loss function*

GateFuseNet is optimized with the binary focal loss [19], which down-weights easy samples so that training emphasizes clinically critical hard cases. Given logits $z \in R^{N \times 1}$ and ground-truth labels $y \in \{0,1\}^{N \times 1}$, the predicted probability is $\hat{p} = \sigma(z)$. The loss is defined as:

$$L_{focal} = -\alpha(1-\hat{p})^{\gamma} y \odot \log \hat{p} - (1-\alpha)\hat{p}^{\gamma}(1-y) \odot \log(1-\hat{p}) \quad (6)$$

with the focusing parameter $\gamma=2.0$ and class-balancing term $\alpha=0.5$, which are selected based on empirically.

*I. Data preprocessing and Network Training*

The dataset was initially divided using an 8:2 split to create training and testing sets, with the training portion further subjected to five-fold cross-validation to ensure robust model evaluation.

Prior to training, all input modalities, including QSM, T1w images, and ROI masks, were first resampled to $1 \times 1 \times 1$ mm³ isotropic resolution to standardize voxel spacing. Each volume was then center-cropped or zero-padded to a uniform size of 128×128×128, ensuring shape compatibility. Real-time data augmentation was implemented including: (i) random affine transformations with rotation angles sampled from [-5°, +5°], translation from [-2, +2] voxels, and scaling factors from [0.9, 1.1] (p=0.2); (ii) random bias field correction (coefficients=0.3, p=0.1); and (iii) additive Gaussian noise ($\sigma=0.02$, p=0.1).

All network in this work were trained with two NVIDIA Tesla V100 (32 GB) GPUs, using Pytorch version of 2.3.0. The networks were optimized using AdamW optimizer with a batch size of 8 for 30 epochs. The learning rate was set as $2 \times 10^{-4}$ and followed by a cosine-annealing schedule that decayed to $1 \times 10^{-7}$ over 30 epochs. For each cross-validation fold, the model with the highest combined validation AUC and F1 score was retained to ensure balanced performance across evaluation metrics.

*J. Evaluation metrics*

The proposed GateFuse Network was compared with ResNeXt [20], AG_SE_ResNeXt [11], and Densformer-MoE [21] in this work using the following quantitative metrics, including accuracy, precision, recall, F1-score, specificity, positive predictive value (PPV), negative predictive value (NPV), area under the ROC curve (AUC) and precision-recall (PR). Receiver operating characteristic (ROC) curves, precision-recall (PR) curves, and confusion matrices were also reported for comparisons of the proposed method with several existing stage-of-the-art methods.

## III. RESULTS

*A. Comparison with Other Competing Methods*

We evaluated GateFuseNet against three representative baselines, using the same three-modality input setting (QSM, T1w, and pathological-aware ROI). As shown in Table 1, GateFuseNet outperformed all competing models across nearly all evaluation metrics, achieving 85.00% accuracy, 84.98% precision, 86.06% recall, 85.48% F1-score, 83.87% specificity, 92.06% AUC, and 92.27% AUPR.

Compared to the weakest baseline (ResNeXt: AUC = 0.8594, AUPR = 0.8571), GateFuseNet achieved substantial improvements: +6.12% in AUC, +6.56% in AUPR, +8.44% in accuracy, +13.33% in recall, and +9.51% in F1-score. Compared to AG_SE_ResNeXt, which incorporates channel-wise attention, our model yielded consistent gains of +3.75% in AUC, +2.55% in AUPR, and +8.12% in accuracy. Although DenseFormer-MoE leverages Transformer-based mechanisms and achieved the highest recall (88.46%) and NPV (90.63%), GateFuseNet still outperformed it in accuracy (+3.75%), F1-score (+6.17%), AUC (+1.22%), and especially AUPR (+9.76%), while also maintaining lower computational complexity and faster inference.

*B. Ablation Study*

To validate the effectiveness of our proposed components, we conduct comprehensive ablation experiments examining four critical aspects: multimodal advantages and position of fusion blocks.

TABLE 1. PERFORMANCE EVALUATION OF THE PROPOSED GATEFUSENET AGAINST THE OTHER THREE PD DIAGNOSING METHODS.

| Model | Accuracy | Precision | Recall | F1_score | Specificity | PPV | NPV | AUC | PR |
|---|---|---|---|---|---|---|---|---|---|
| ResNeXt | 0.7656 | 0.8063 | 0.7273 | 0.7597 | 0.8065 | 0.8063 | 0.7404 | 0.8594 | 0.8571 |
| AG_SE_ResNeXt | 0.7688 | 0.7996 | 0.7576 | 0.7705 | 0.7806 | 0.7996 | 0.7574 | 0.8831 | 0.8972 |
| Densformer-MoE | 0.8125 | 0.7188 | **0.8846** | 0.7931 | 0.7632 | 0.7188 | **0.9063** | 0.9084 | 0.8251 |
| GateFuseNet | **0.8500** | **0.8498** | 0.8606 | **0.8548** | 0.8387 | **0.8498** | 0.8520 | **0.9206** | **0.9227** |

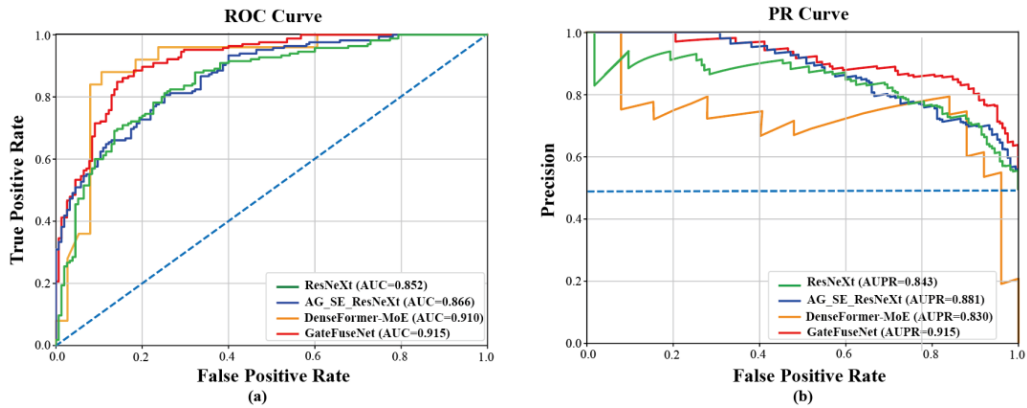

Fig. 2 ROC (a) and PR (b) curve analysis for Parkinson's disease diagnosis performance comparison.

**Fusion Methodology Effectiveness:** We evaluate the effectiveness of our proposed GF blocks against conventional fusion strategies. As shown in Table 2, weighted average fusion achieves 76.68% accuracy and 86.42% AUC, while concatenation-based fusion demonstrates improved performance with 78.17% accuracy and 88.23% AUC. The propsoed approach outperforms both baseline methods, achieving 85.00% accuracy and 92.06% AUC. The improvements of 6.83% in accuracy and 3.83% in AUC compared to concatenation fusion demonstrate the effectiveness of our attention-driven multimodal integration strategy.

TABLE 2. EFFECTIVENESS OF THE PROPOSED FUSION MODULE.

| Fusion method | Accuracy | Precision | Recall | F1-score | AUC |
|---|---|---|---|---|---|
| Weighted averaging | 0.7668 | 0.8057 | 0.7818 | 0.7983 | 0.8642 |
| Simple concatenation | 0.7817 | 0.8165 | 0.7939 | 0.8099 | 0.8823 |
| The propsoed | **0.8500** | **0.8498** | **0.8606** | **0.8548** | **0.9206** |

**Effects of Fusion Block Placement:** We investigate the influence of fusion block placement across different input branches to identify the most effective fusion strategy. As shown in Table 3, placing GF blocks in the T1w and QSM branches yielded accuracies of 77.06% and 78.43%, respectively. In contrast, positioning the fusion module within the ROI branch resulted in the highest performance, achieving 85.00% accuracy and 92.06% AUC. These findings support our hypothesis that the pathological-aware ROI branch provides a more informative and anatomically grounded feature stream, making it a more suitable anchor point for multimodal integration in PD diagnosis.

TABLE 3. INVESTIGATION ON THE POSITIONS OF THE PROPOSED FUSION MODULE.

| Effects of Fusion Block Placement | Accuracy | Precision | Recall | F1 score | AUC |
|---|---|---|---|---|---|
| T1 branch | 0.7706 | 0.7983 | 0.7436 | 0.7647 | 0.8761 |
| QSM branch | 0.7843 | 0.8148 | 0.7576 | 0.7822 | 0.8920 |
| ROI branch | **0.8500** | **0.8498** | **0.8606** | **0.8548** | **0.9206** |

## C. Model interpretation

To demonstrate the interpretability and clinical relevance of our proposed GateFuseNet, we visualize the learned feature regions contributing most significantly to PD diagnosis. Fig.4 presents the fused feature activation maps overlaid on representative axial slices, with the leftmost panel showing anatomical reference landmarks including the CN, PU, GP, SN, and RN.

The Grad-CAM heatmaps reveal that GateFuseNet consistently attends to clinically relevant deep DGM structures, with particularly strong activations in the GP and SN. The red-yellow colormap highlights regions of high discriminative importance, predominantly within the basal ganglia, which closely corresponds to established Parkinson's disease pathology, suggesting that the ROI-guided adaptive fusion mechanism effectively captures pathological cues such as iron deposition and structural changes associated with PD.

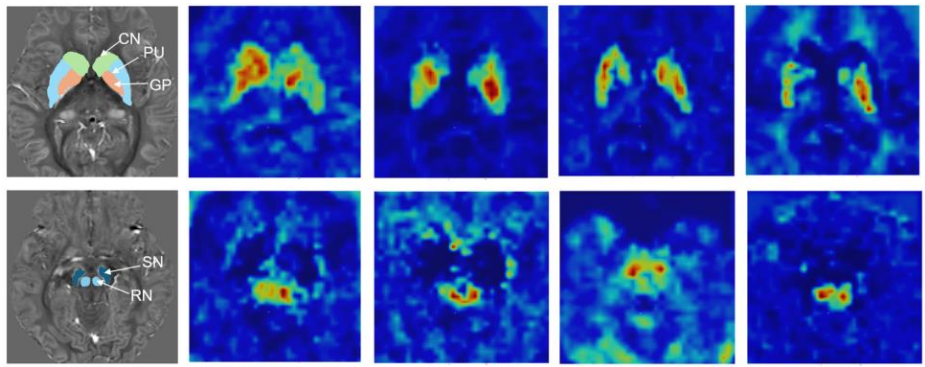

Fig. 3 Grad-CAM heatmaps generated by GateFuseNet on different axial slices of testing-set samples from four patients with PD.

## IV. Discussion and Conclusions

In this study, we proposed GateFuseNet, an ROI-guided multimodal fusion framework that enhances PD diagnosis by jointly leveraging QSM, T1-weighted imaging, and pathological ROI priors. The architecture integrates complementary structural and pathological information through proposed Gated Fusion (GF) mechanisms and attention, selectively emphasizing clinically relevant features. Comprehensive comparisons with state-of-the-art methods (ResNeXt [25], AG-SE-ResNeXt [15], and DenseFormer-MoE [26]) demonstrated consistent superiority across all evaluation metrics.

Ablation analyses further validated the diagnostic value of key components. Assigning the pathological ROI as the primary branch yielded the most significant gains, emphasizing regions directly implicated in PD-related degeneration. The multimodal integration improved discriminative power, while the proposed GF mechanism outperformed conventional fusion strategies by enabling dynamic, attention-guided feature modulation. These findings underscore the benefit of anatomically informed multimodal design for robust and interpretable PD diagnosis.

Importantly, Grad-CAM visualizations confirmed that GateFuseNet consistently focuses on pathologically relevant deep gray matter regions (e.g., SN, GP), aligning well with known sites of dopaminergic loss and iron deposition. This correspondence indicates that the model captures physiologically meaningful cues rather than relying on spurious correlations, reinforcing its clinical interpretability.

Despite promising results, this study has several limitations. The single-center dataset may limit generalizability, highlighting the need for multi-site validation and domain adaptation. Future extensions could incorporate additional modalities such as diffusion or functional MRI and integrate longitudinal imaging with clinical scores to enhance prognostic and personalized applications.

## Acknowledgment

This work was supported by the National Natural Science Foundation of China under grant number 62301616 and the Natural Science Foundation of Hunan under grant number 2024JJ6530，and the support from the Australia Research Council (DE20101297 and DP230101628).